\pdfoutput=1

\documentclass[11pt]{article}

\usepackage[table,xcdraw]{xcolor}
\usepackage[final]{acl}

\usepackage{times}
\usepackage{latexsym}

\usepackage[T1]{fontenc}

\usepackage[utf8]{inputenc}

\usepackage{microtype}

\usepackage{inconsolata}

\usepackage{graphicx}

\usepackage{booktabs}
\usepackage{arydshln}
\usepackage{amsmath}
\usepackage{amssymb}
\usepackage[most]{tcolorbox}
\usepackage{makecell}
\usepackage{enumitem}
\usepackage{mathtools}

\usepackage{hhline}
\usepackage{tabularx}
\usepackage{multirow}
\usepackage{mathrsfs}
\usepackage{tabulary}
\usepackage[ruled,linesnumbered]{algorithm2e}
\usepackage{amsfonts,amssymb}
\usepackage{amsthm}
\usepackage{wrapfig}
\usepackage{listings}
\usepackage{algorithmic} 
\usepackage{color, colortbl}
\definecolor{ll1}{HTML}{fffadc}
\definecolor{ghlightbackground}{HTML}{FFFFFF}
\definecolor{ghlightcomment}{rgb}{0,0.6,0}
\definecolor{ghlightkeyword}{HTML}{d0212f}
\definecolor{ghlightstring}{HTML}{032F62}
\definecolor{ghlightnumber}{HTML}{005CC5}

\lstdefinestyle{mystyle}{
    language=Python,
    backgroundcolor=\color{ghlightbackground},
    commentstyle=\color{ghlightcomment},
    keywordstyle=\color{ghlightkeyword},
    numberstyle=\tiny\color{ghlightnumber},
    stringstyle=\color{ghlightstring},
    basicstyle=\ttfamily\scriptsize,
    columns=fullflexible,
    breaklines=true,
    captionpos=b,
    escapechar=|,
    breakatwhitespace=false,
    keepspaces=true,
    numbersep=5pt,
    showspaces=false,
    showstringspaces=false,
    showtabs=false,
    tabsize=2,
}
\makeatletter
  \let\orig@hdashline\hdashline
  \renewcommand{\hdashline}{%
    \noalign{\vskip 0.15em}
    \orig@hdashline
    \noalign{\vskip 0.15em}
  }
\makeatother

\newtcolorbox[list inside=prompt,auto counter,number within=section]{prompt}[1][]{
    colbacktitle=black!60,
    coltitle=white,
    fontupper=\ttfamily\footnotesize,
    boxsep=5pt,
    left=0pt,
    right=0pt,
    top=0pt,
    bottom=0pt,
    boxrule=1pt,
    width=\textwidth,
    #1,
}

\title{Self-Correcting Code Generation Using Small Language Models}
\author{
Jeonghun Cho\textsuperscript{1},
Deokhyung Kang\textsuperscript{1},
Hyounghun Kim\textsuperscript{1,2},
Gary Geunbae Lee\textsuperscript{1,2}
\\
\textsuperscript{1}Graduate School of Artificial Intelligence, POSTECH\\
\textsuperscript{2}Department of Computer Science and Engineering, POSTECH\\
\texttt{\{jeonghuncho, deokhk, h.kim, gblee\}@postech.ac.kr}
}

\begin{document}
\maketitle

\begin{abstract}
Self-correction has demonstrated potential in code generation by allowing language models to revise and improve their outputs through successive refinement. Recent studies have explored prompting-based strategies that incorporate verification or feedback loops using proprietary models, as well as training-based methods that leverage their strong reasoning capabilities. However, whether smaller models possess the capacity to effectively guide their outputs through self-reflection remains unexplored. Our findings reveal that smaller models struggle to exhibit reflective revision behavior across both self-correction paradigms. In response, we introduce \textbf{\textsc{CoCoS}}, an approach designed to enhance the ability of small language models for multi-turn code correction. Specifically, we propose an online reinforcement learning objective that trains the model to confidently maintain correct outputs while progressively correcting incorrect outputs as turns proceed. Our approach features an accumulated reward function that aggregates rewards across the entire trajectory and a fine-grained reward better suited to multi-turn correction scenarios. This facilitates the model in enhancing initial response quality while achieving substantial improvements through self-correction. With 1B-scale models, \textsc{CoCoS} achieves improvements of 35.8\% on the MBPP and 27.7\% on HumanEval compared to the baselines.\footnote{Our implementation can be accessed at \url{https://github.com/jeonghun3572/CoCoS}}
\end{abstract}
\section{Introduction}
Despite the progress in large language models (LLMs), generating the correct code snippets in a single attempt remains a challenging task across many programming scenarios. In response, recent studies have explored the use of external supervision, typically provided by a more capable teacher model, to guide the correction process~\citep{welleck2023generating, yang2025supercorrect}. These approaches typically leverage teacher models, which either guide student models through iterative self-correction via feedback or perform both generation and correction in a self-directed manner without external models~\citep{kamoi-etal-2024-llms}.

Given the dependence on powerful guidance, most of the successful results have been achieved using strong proprietary models such as Gemini~\citep{team2023gemini,team2024gemini} and GPT-series~\citep{NEURIPS2020_1457c0d6,gpt4.1}, which are only available through external APIs~\citep{chen2024teaching, chen2025sets}. Considering the reliance on costly proprietary models in previous studies, it remains uncertain whether smaller, open-source language models (SLMs) can effectively guide their outputs through self-reflection~\citep{huang2024large,Han_Liang_Shi_He_Xiao_2024}.
To investigate this, we explore whether SLMs can achieve meaningful self-correction without proprietary systems. Our findings show that smaller models struggle to exhibit reflective revision behavior when they rely solely on prompting (\S\ref{sec:prompting result}). Building on these observations, we aim to explore training-based self-correction for code generation. However, we find that such methods are not directly applicable to SLMs, likely due to their assumption of sufficient self-correction capability (\S\ref{sec:score}).

To tackle these limitations, we introduce \textbf{\textsc{CoCoS}} (Self-\textbf{Co}rrecting \textbf{Co}de generation using \textbf{S}mall LMs), a reinforcement learning (RL) framework designed to support intrinsic self-correction in small LMs. Our approach leverages an accumulated reward function that considers previous responses and outputs a scalar reward that captures the cumulative effect of multiple turns. This function encourages both successful self-correction and improved initial response quality by rewarding the entire response trajectory. Additionally, our method adopts a progressive reward that evaluates incremental improvements in code quality by tailoring the setting to a multi-turn scenario, which offers a more fine-grained assessment than binary rewards based solely on code correctness. We validate our approach on 1B-scale models across various code generation tasks and observe robust generalization in unseen settings.

Our contributions are as follows: (1) We empirically demonstrate that previous prompting-based and training-based self-correction methods struggle to generalize to small language models.
(2) We introduce a reinforcement learning reward function tailored for the multi-turn code generation setting, leveraging accumulated rewards with a discount factor and fine-grained assessments to encourage effective self-correction.
(3) We validate our approach across diverse Python code generation datasets and show consistent improvements over baselines, including in unseen settings.
\section{Related Work}
\paragraph{Prompting-based correction guidance.}
Several recent studies on self-correction have leveraged the advanced reasoning capabilities of LLMs to improve the accuracy of their initial responses~\citep{zelikman2022star,renze2024self}. In coding tasks that require complex reasoning, self-correction often involves reviewing generated code snippets to identify and fix errors with the model's own judgment. However, prior works suggest that naive prompting for self-correction may reduce performance~\citep{huang2024large, NEURIPS2024_639d992f}. As a result, SETS~\citep{chen2025sets} combines sampling, self-verification, and self-correction into a unified framework that facilitates LLMs to improve their own outputs. Self-Refine~\citep{NEURIPS2023_91edff07} conducts iterative evaluation and correction of the initial responses using few-shot prompting without any additional fine-tuning.
However, these methods rely on the assumption that the model is already capable of revising its outputs.

\paragraph{Training-based correction guidance.}
In addition to the above methods, fine-tuning has been explored to endow models with intrinsic self-correction.
Self-Corrector~\citep{welleck2023generating} employs a separate correction model that shares the same backbone as the initial response generator. \citet{zhang-etal-2024-small} introduces a distinct verifier model to guide the revision process. \citet{yang2025supercorrect} propose a teacher-guided framework where a stronger model supervises both the intermediate reasoning steps and the final prediction of an SLM. However, all of these methods depend on training separate corrector or verifier models, which require additional computational resources.

\paragraph{Self-correction without external guidance.}
ReVISE~\citep{lee2025revise} implements internal self-correction by generating special tokens that determine whether to revise or terminate the current output. SCoRe~\citep{kumar2025training} adopts a two-stage RL framework that provides rewards to both the previous and current responses to guide learning.
While both methods have demonstrated improved code generation, ReVISE has limited generalization beyond the training turns due to its reliance on supervised fine-tuning (SFT)~\citep{chu2025sft}. SCoRe is the first work to alleviate this limitation through online learning with self-generated data. While both SCoRe and our approach leverage online-RL, SCoRe remains constrained to proprietary models, whereas \textsc{CoCoS} demonstrates effectiveness even for SLMs that struggle with self-correction.
\section{Problem Description}\label{sec:problem description}
In this work, we focus on code generation tasks and study a small-scale LLM to investigate its intrinsic ability to improve its outputs over subsequent trials. Suppose $\mathcal{D}=\{(x, y, u)\}$ consists of data points, each represented as a tuple $(x,y,u)$, where $x$ is a problem, $y$ is the canonical code snippet, and $u$ is the unit test cases. We define the SLM as $\pi_\theta$ and aim to generate correct $\hat{y}_t$ through the conditional distribution $\pi_\theta(\hat{y}_t | x, \hat{y}_{1:t-1}, p_{1:t-1})$, where $\hat{y}_{1:t-1}$ represents the history of SLM's previous trials, and $p_{1:t-1}$ denotes auxiliary instructions for revising the previous responses. Note that the subscripts indicate turn indices within the multi-turn setting. Since we do not consider external feedback (e.g., compiler outcomes, other LMs' feedback), $p_{1:t-1}$ is fixed for each turn.
\begin{table*}[t]
\centering
\begin{tabular}{lccccc}
\toprule
\textbf{Method} & \textbf{Accuracy@t1} & \textbf{Accuracy@t2} & $\Delta^{\mathrm{i} \rightarrow \mathrm{c}}$\textbf{(t1, t2)} & $\Delta^{\mathrm{c} \rightarrow \mathrm{i}}$\textbf{(t1, t2)}$\downarrow$ & \textbf{$\Delta^{\text{Acc}}$(t1, t2)} \\ \midrule
Simple-prompting & 43.8\% & 44.6\% & 1.4\% & 0.6\% & 0.8\% \\ \hdashline
Self-Refine & 43.8\% & 38.2\% & 2.8\% & 8.4\% & $-$5.6\% \\
External-Refine & 43.8\% & 41.8\% & 7.0\% & 9.0\% & $-$2.0\% \\
SETS & 39.3\% & 38.8\% & - & - & - \\
External-SETS & 39.3\% & \textbf{49.8\%} & - & - & - \\
\bottomrule
\end{tabular}
\caption{Performance comparison of prompting-based methods.}
\label{tab:prompting-based}
\end{table*}
\section{Preliminary Study: Prompting-based Self-Correction}\label{sec:prompting-based}
Prompting-based self-correction typically involves incorporating feedback into the prompt in the form of the auxiliary instruction sequence $p_{1:t-1}$~\citep{kamoi-etal-2024-llms}.
This feedback can be either provided by a separate teacher model or generated by the model itself. In this section, we examine both settings to assess whether prompting-based self-correction is effective for small language models.

\subsection{Metrics}
To evaluate self-correction, we follow the metrics used in \citet{kumar2025training}.
An output $y_t$ is considered correct if it passes all test cases and is incorrect otherwise. \textbf{Accuracy@t1} and \textbf{Accuracy@t2} denote the accuracy at the first and second turns, respectively.
\textbf{$\Delta^{\mathrm{i} \rightarrow \mathrm{c}}$(t1, t2)} denotes the fraction of problems that change from incorrect to correct between the first and second turns, and \textbf{$\Delta^{\mathrm{c} \rightarrow \mathrm{i}}$(t1, t2)} vice versa.
Finally, \textbf{$\Delta^\textbf{Acc}$(t1, t2)} measures the change in accuracy between the first and second turns.

\subsection{Setup}
We experiment with two prompting-based self-correction methods, \textbf{Self-Refine}~\citep{NEURIPS2023_91edff07} and \textbf{SETS}~\citep{chen2025sets}—both demonstrated exclusively on proprietary LLMs in their original work.
To verify the effectiveness of SLMs, all experiments are conducted using the pre-trained Qwen2.5-1.5B~\citep{yang2024qwen2}. 
Self-Refine uses 3-shot prompting, whereas SETS relies on test-time scaling. Since SETS does not generate a fixed initial response, we estimate Accuracy@t1 by averaging the accuracy of 20 sampled initial outputs. Accuracy@t2 is similarly computed after all correction steps. However, because SETS verifies each sample and only applies correction to those deemed correct, $\Delta$ improvements between initial and final responses cannot be consistently measured. Instead, we report feedback accuracy in Appendix~\ref{apdx:sets}.

We further investigate the influence of external feedback by employing strong teacher models,\footnote{\label{foot:gpt4.1}We use gpt-4.1-2025-04-14~\citep{gpt4.1} for teacher model.} which we refer to as \textbf{External-Refine} and \textbf{External-SETS}. For comparison, we also introduce a simple correction prompting method that first generates an initial response with a 3-shot prompt, then prompts for correction without disclosing the correctness of the initial output. Additional details including prompts and hyperparameters are provided in Appendix~\ref{apdx:prompts}.

\subsection{Results}\label{sec:prompting result}
In conclusion, SLMs struggle to revise their responses using prompting alone, as shown in Table~\ref{tab:prompting-based}. Although simple prompting yields a marginal performance gain ($\Delta^{\text{Acc}}$ of 0.8), it fails to enable meaningful revisions. Both prior works, when used without an external LLM for self-verification, underperform compared to simple prompting and frequently modify responses that were initially correct. Effective self-correction becomes only possible when a strong external LLM provides feedback (External-SETS; 10.5\% gain in Accuracy@t1). However, this setup assumes access to a powerful external model\textsuperscript{~\ref{foot:gpt4.1}} and requires calling it alongside the SLM during inference, which is impractical in real-world scenarios.
\begin{figure*}[t]
\centering
\includegraphics[width=\textwidth,keepaspectratio]{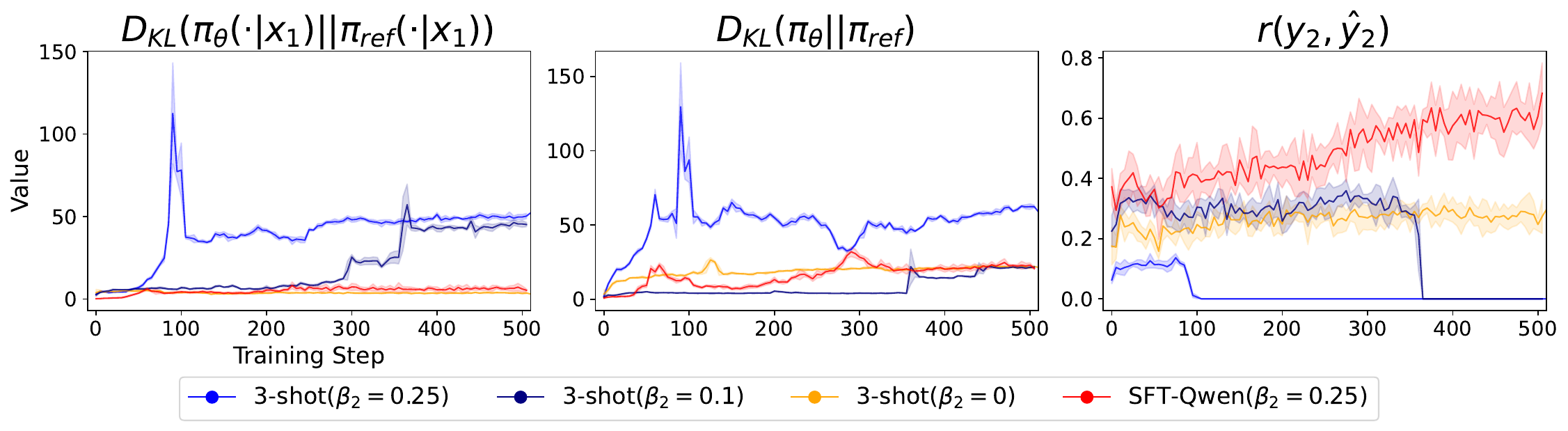}
\caption{Learning curves for SCoRe based on experiments with Qwen2.5-1.5B. \textbf{(1)} KL-regularization for the first turn, \textbf{(2)} Default KL-divergence penalty for the policy gradient training, and \textbf{(3)} Reward for the second response. \textbf{\textit{The KL-regularization for the first turn becomes excessively high as the training step increases so that the training collapses.}} We also observe collapse in another small model, which we analyze in Appendix~\ref{apdx:score}.}
\label{fig:score reward}
\end{figure*}

\section{Proposed Methodology}\label{sec:proposed methodology}
Based on the observations in \S\ref{sec:prompting-based}, SLMs struggle to demonstrate reflective revision behavior when relying solely on prompting, without access to external LLMs. In this section, we introduce a training-based method designed to enable effective self-correction.

\subsection{Multi-turn MDP}
As outlined in \S\ref{sec:problem description}, our goal is to enhance model outputs across multiple turns by leveraging the response history. Since the model's own generated history $\hat{y}_{1:t-1}$ naturally defines a sequential decision process, we formulate the process of self-correction as a multi-turn Markov decision process~\citep{NEURIPS2024_639d992f}, where the policy is defined as
\begin{equation}\nonumber
    \pi_\theta(\hat{y}_t \mid x, \hat{y}_{1:t-1}, p_{1:t-1})
\end{equation}
In this formulation, the state at each turn $t$ is denoted as $s_t = (x, \hat{y}_{1:t-1}, p_{1:t-1})$, which consists of the input $x$, the model's prior outputs $\hat{y}_{1:t-1}$, and the prompt history $p_{1:t-1}$. The action $a_t$ corresponds to generating a revised output $\hat{y}_t$. This action is then appended to form the next prompt $p_t$ for the subsequent turn.
Accordingly, the policy can be rewritten as $\pi_\theta(a_t \mid s_t)$. We evaluate the correctness of the generated code $\hat{y}_t$ using unit test execution $u(\hat{y}_t)$, which assesses the functional correctness of the code against the given unit tests. The test outcome is used as the reward:
\begin{equation}\nonumber
    r(y_t, \hat{y}_t) \coloneqq u(\hat{y}_t)
\end{equation}
where $r(y_t, \hat{y}_t)$ denotes the reward for $\hat{y}_t$.

Under this formulation, the learning objective is to optimize a policy $\pi_\theta$ that maximizes the expected cumulative reward over a refinement trajectory:
\begin{equation}\nonumber
\begin{split}
    \max_{\theta} \; & \mathbb{E}_{(x,y) \sim \mathcal{D},\; \hat{y}\ \sim \pi_\theta} \Big[\sum_{t=1}^{T} r(y_t, \hat{y}_t) \\
    & - \beta \mathbb{D}_{KL}\big(\pi_\theta (\cdot \mid s_t) \;\|\; \pi_{ref}(\cdot \mid s_t)\big)\Big]
\end{split}
\end{equation}
where $\beta > 0$ is the Kullback-Leibler (KL) coefficient that controls the strength of regularization against the reference policy $\pi_{ref}$.

\subsection{Observed limitations}\label{sec:score}
\begin{figure}[t]
\centering
\includegraphics[width=\columnwidth,keepaspectratio]{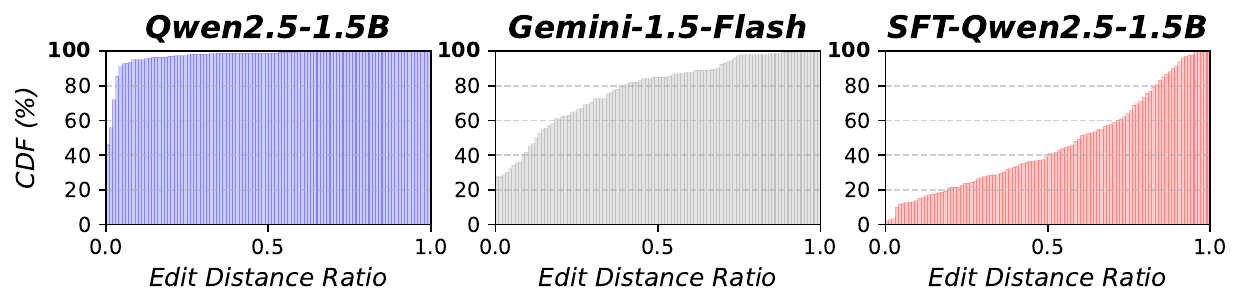}
\caption{Cumulative distribution functions (CDFs) of edit distance ratios between the first and second responses, evaluated on Qwen2.5-1.5B and Gemini-1.5-Flash under the 3-shot setting.}
\label{fig:edit_distance}
\end{figure}
Previous training-based studies have used supervised learning to pair incorrect responses with corrections, enabling the model to revise incorrect outputs for multi-turn self-correction~\citep{zelikman2022star,welleck2023generating}.
However, in an ideal scenario, LLMs also should aim to complete the given task on the first attempt, minimizing the need for correction. In other words, the goal of self-correction learning is to maximize not only Accuracy@t2 but also Accuracy@t1. To this end, SCoRe~\citep{kumar2025training} proposed a two-stage approach to optimize both turns. Specifically, in the first stage, training aims to maximize the reward of the second turn while applying strong KL-regularization to the first turn to preserve the distribution of the initial response. In the second stage, the model jointly optimizes both responses.
This approach is the first to achieve positive intrinsic self-correction. The training formulation in the first stage is as follows:
\begin{equation}\label{eq:score}
\begin{split}
    r(y_2, \hat{y}_2) & - \beta \sum_{t=1}^{2}\mathbb{D}_{KL} \big(\pi_\theta(\cdot \mid s_t) \;\|\; \pi_{ref}(\cdot \mid s_t)\big) \\
    &  -\beta_2 \mathbb{D}_{KL}\big(\pi_\theta(\hat{y}_1\mid s_1) \;\|\; \pi_{ref}(y_{ref}\mid s_1)\big)
\end{split}
\end{equation}
where $r(y_2, \hat{y}_2)$ is a binary reward, taking a value of $1$ for all-pass and $0$ for fail. The second KL-divergence term represents the regularization, and $\beta \ll \beta_2$. However, we find that introducing the second KL-divergence term ($\beta_2>0$) leads to training collapse for SLMs, as shown in Figure~\ref{fig:score reward} where the reward converges to zero.

We hypothesize that training collapse arises from the limited self-correction ability of SLMs. In the simple-prompting baseline (Table~\ref{tab:prompting-based}), the low values of $\Delta^{\mathrm{i} \rightarrow \mathrm{c}}$ and $\Delta^{\mathrm{c} \rightarrow \mathrm{i}}$ suggest that the model rarely revises its initial response, often producing near-identical outputs across turns ($\hat{y}_1 \approx \hat{y}_2$). This is further supported by Figure~\ref{fig:edit_distance}, where Qwen2.5-1.5B shows minimal changes (edit distance $\le 0.05$) in 93\% of cases when the second response is incorrect, compared to only 32\% for Gemini.

This lack of revision becomes problematic when optimizing the reward objective in Equation~\ref{eq:score}. While the reward term encourages updates to $y_2$, the strong overlap with $y_1$ causes the KL regularization—applied to $y_1$—to inadvertently constrain $y_2$, thereby destabilizing training and limiting reward-driven updates.

To further investigate our hypothesis, we construct an SFT-Qwen model that is intentionally fine-tuned to produce outputs where $y_1 \neq y_2$. The edit distance between $y_1$ and $y_2$ can be observed in the right histogram in Figure~\ref{fig:edit_distance}. To this end, we create training data by mixing $\hat{y}_1$ and $\hat{y}_2$ generated from Pre-trained Qwen and Gemini, respectively, and deliberately overfit the model to MBPP to instill correction capability.
Figure~\ref{fig:score reward} illustrates the impact of KL-regularization on Pre-trained Qwen and SFT-Qwen. We observe that first-turn KL-regularization explodes during training. This leads to instability in the learning process, driving the reward toward zero and ultimately resulting in training collapse. In contrast, disabling this regularization (i.e., $\beta_2 = 0$) or using SFT-Qwen prevents such instability.
Interestingly, SFT-Qwen does not exhibit training collapse, even when trained with a KL-regularization coefficient ($\beta_2 = 0.25$)—a large value that reflects SCoRe's intention—thus supporting our hypothesis about the conditions that give rise to instability. Building on this experiment, we propose a method to enhance intrinsic self-correction even in SLMs.

\subsection{\textsc{CoCoS}}
To enable self-correction, it is crucial to capture the difference between the initial and subsequent responses. A straightforward approach—maximizing only the difference of rewards at each turn by subtracting $r(y_1,\hat{y}_1)$ from $r(y_2,\hat{y}_2)$—may lead to reward hacking, where the model intentionally degrades $r(y_1,\hat{y}_1)$ to inflate the reward difference~\citep{10.5555/3600270.3600957}. Therefore, SCoRE introduced the KL-regularization to prevent this collapse; however, SLMs struggle with this strategy (\S\ref{sec:score}). To address this, we introduce \textsc{CoCoS}, an RL-based method that uses a discount factor $\gamma$ to accumulate rewards from prior responses, preventing the collapse of the initial response while also providing fine-grained assessments through progressive reward that capture incremental improvements between turns.

Since our setting is also formulated as a multi-turn MDP, we adopt a REINFORCE-style policy gradient method~\citep{ahmadian-etal-2024-back} to optimize the policy, which is widely used in single-turn RLHF ($\gamma=0$). In order to adapt to a multi-turn setting, we introduce two key modifications to the reward function. First, we introduce an accumulated reward function that incorporates a discount factor $\gamma$ to encourage the model to consider the assessment of the previous turn; in the case of SCoRe, $\gamma=0$ since it computes the reward independently at each turn. Unlike conventional RL, where $\gamma$ discounts future rewards, we adapt its usage to amplify the contribution of the most recent turn. The accumulated reward function $R(\hat{y}_{1:T})$ is defined as follows (with $T=2$ in our setup):
\begin{equation}\label{eq:reward function}
R(\hat{y}_{1:T}) = \gamma^{T-1} r_1 + \sum_{t=2}^T \gamma^{T-t} (r_t - r_{t-1})
\end{equation}
where $r_t$ denotes the reward at the $t$-th turn. When $\gamma<1$, the function focuses more on recent changes in reward, whereas $\gamma>1$, in principle, places greater emphasis on earlier responses. Based on empirical observations, we set $\gamma=0.5$ in our experiments. The impact of varying $\gamma$ is discussed in \S\ref{sec:discount factor}.

In multi-turn settings, capturing the model’s progress over successive turns is important. However, prior work has typically used a binary reward that reflects whether all test cases pass, which makes it difficult to account for gradual improvements during the refinement process~\citep{zheng2025what}. To address this, we introduce a progressive reward based on the pass ratio. We assume that each code is paired with $K$ unit test cases to assess semantic correctness, and the progressive reward is computed as follows:
\begin{equation}\label{eq:progressive reward}
r_t = \frac{1}{K}\sum_{k=1}^K \mathbb{I}\{ u_k(\hat{y}_t) \text{ passes} \}
\end{equation}
The effectiveness of the progressive reward compared to the binary scheme is presented in \S\ref{sec:progressive reward}. Accordingly, our training objective is defined as follows:
\begin{equation}\label{eq:cocos}
\begin{split}
    \max_{\theta} \; & \mathbb{E}_{(x,y) \sim \mathcal{D}, \; \hat{y}\ \sim \pi_\theta} \Big[R(\hat{y}_{1:T}) \\
    &- \beta \sum_{t=1}^{T}\mathbb{D}_{KL}\big(\pi_\theta(\cdot \mid s_t) \;\|\; \pi_{ref}(\cdot \mid s_t)\big)\Big]
\end{split}
\end{equation}
where $\beta$ denotes the KL coefficient. Further experimental details can be found in Appendix~\ref{apdx:detail}.
\begin{table}[t]
\centering
\resizebox{\columnwidth}{!}{%
\begin{tabular}{lcc}
\toprule
\textbf{Method} & \textbf{Accuracy@t1} & \textbf{Accuracy@t2} \\ \midrule
\multicolumn{3}{l}{\textbf{Pre-Trained (3-shot)}} \\
Qwen2.5-1.5B & 43.8\% & 44.6\% \\
Llama-3.2-1B & 27.6\% & 27.6\% \\
deepseek-coder-1.3B & 46.4\% & 46.0\% \\ \hdashline
\multicolumn{3}{l}{\textbf{Boost model (0-shot)}} \\
Qwen2.5-1.5B & 9.0\% \textcolor{red}{(34.8\%$\downarrow$)} & 10.2\% \textcolor{red}{(34.4\%$\downarrow$)} \\
Llama-3.2-1B & 10.4\% \textcolor{red}{(17.2\%$\downarrow$)} & 11.0\% \textcolor{red}{(16.6\%$\downarrow$)} \\
deepseek-coder-1.3B & 16.6\% \textcolor{red}{(29.8\%$\downarrow$)} & 20.2\% \textcolor{red}{(25.8\%$\downarrow$)} \\ \bottomrule
\end{tabular}%
}
\caption{Performance comparison between pre-trained and boost models on the MBPP dataset. Pre-Trained models are evaluated using 3-shot prompting.
\textbf{\textit{The results suggest that using boost models as the backbone in our experiments has minimal impact.}}}
\label{tab:backbone}
\end{table}
\begin{table*}[t]
\centering
\begin{tabular}{lccccc}
\toprule
\textbf{Method} & \textbf{Accuracy@t1} & \textbf{Accuracy@t2} & $\Delta^{\mathrm{i} \rightarrow \mathrm{c}}$\textbf{(t1, t2)} & $\Delta^{\mathrm{c} \rightarrow \mathrm{i}}$\textbf{(t1, t2)}$\downarrow$ & \textbf{$\Delta^\textbf{Acc}$(t1, t2)} \\ \midrule
Boost model & 9.0\% & 10.2\% & 1.2\% & 0.0\% & 1.2\% \\
Turn-SFT & 44.2\% & 44.4\% & 1.4\% & 1.2\% & 0.2\% \\
Self-Corrector & 43.8\% & 48.8\% & 11.2\% & 6.2\% & 5.0\% \\
ReVISE & 34.6\% & 41.2\% & 11.3\% & 4.9\% & 6.4\% \\
Turn-RL & \textbf{46.4\%} & 48.6\% & 4.0\% & 1.8\% & 2.2\% \\\hdashline
\textsc{CoCoS} & 45.0\% & \textbf{54.2\%} & 11.0\% & 1.8\% & 9.2\% \\
\bottomrule
\end{tabular}
\caption{Main results of Qwen2.5-1.5B on the MBPP dataset. The highest accuracy values are highlighted in bold. \textbf{\textit{CoCoS has the highest selective correction rate, correcting incorrect responses without changing correct ones. It also demonstrates strong generalization, achieving high correction success rates even on unseen data.}}}
\label{tab:main_result_qwen}
\end{table*}
\section{Experimental Settings}
As prior studies are insufficient for enabling intrinsic self-correction (\S\ref{sec:prompting-based}), we compare \textsc{CoCoS} against fine-tuning approaches. To support this comparison, we introduce the experimental settings, including descriptions of the models (\S\ref{sec:models}), datasets (\S\ref{sec:datasets}), and baselines (\S\ref{sec:baselines}) used in our experiments.

\subsection{Models}\label{sec:models}
We use Qwen2.5-1.5B~\citep{yang2024qwen2}, Llama-3.2-1B~\citep{grattafiori2024llama}, and DeepSeek-Coder-1.3B-Base~\citep{guo2024deepseek} as base models and fine-tune them for each method. During inference, we use greedy decoding to ensure reproducibility. Further implementation details are provided in Appendix~\ref{apdx:detail}.

\subsection{Datasets}\label{sec:datasets}
All models are trained on MBPP~\citep{austin2021program} and evaluated on MBPP, HumanEval~\citep{chen2021evaluating}, and ODEX~\citep{wang-etal-2023-execution} to assess generalization to unseen data. For MBPP, we follow the data split provided in \citet{austin2021program}, which defines train, validation, test, and few-shot sets. We do not use the validation set for model selection; instead, we select the checkpoint that achieves the highest reward during training. Also, few-shot data is excluded.

While supervised fine-tuning (SFT) allows models to learn proper code formatting through input-output pairs, online-RL lacks such structural guidance. Therefore, we perform SFT on the KodCode~\citep{xu2025kodcode} dataset to produce parseable code so that RL training can be conducted with unit test execution.
KodCode is curated to exclude data from MBPP, HumanEval, and ODEX. We measure the model's accuracy on MBPP at both the pre-SFT and post-SFT, which we denote as the \textbf{Pre-trained} and \textbf{Boost model}, respectively, and report the results in Table~\ref{tab:backbone}. As shown in Table~\ref{tab:backbone}, we emphasize that the purpose of SFT on KodCode is not to enhance performance on the target dataset. The \textbf{Boost model} is the backbone for both \textsc{CoCoS} and all baseline comparisons.

\subsection{Baselines}\label{sec:baselines}
We compare \textsc{CoCoS} with two training-based methods: \textbf{Self-Corrector}, which adopts a separate corrector model, and \textbf{ReVISE}, which follows an intrinsic self-correction approach. Since ReVISE is trained on a fixed two-turn dataset, it is constrained to generate a terminate token within two turns. This poses a limitation in terms of turn scalability. To enable multi-turn correction beyond two turns, we train our own SFT baseline, \textbf{Turn-SFT}, using fixed 1-turn and 2-turn examples. Turn-SFT can generalize beyond two turns by following the trained data format during inference, offering better turn scalability than ReVISE and making it suitable for analyzing self-correction beyond two turns.

However, such SFT-based methods still rely on the distribution of the training dataset, which highlights the need for RL–based approaches with better generalization. To compare with an RL-based approach, we also train \textbf{Turn-RL}, where the first response is fixed and the second turn is optimized according to Equation~\ref{eq:cocos}. In other words, Turn-RL constrains the optimization of the initial response in contrast to \textsc{CoCoS}. Further details on the baselines are provided in Appendix~\ref{apdx:baseline}.
\section{Main Results}
As shown in Table~\ref{tab:main_result_qwen}, \textsc{CoCoS} demonstrates the highest correction performance across all turns. Although the baselines are trained to actively revise prior responses, the $\Delta^{\mathrm{c} \rightarrow \mathrm{i}}$ metric indicates that they frequently make unnecessary revisions to already correct outputs. \textsc{CoCoS}, on the other hand, corrects selectively: it achieves a low $\Delta^{\mathrm{c} \rightarrow \mathrm{i}}$ rate of 1.8\%, avoiding unnecessary changes, while maintaining a high $\Delta^{\mathrm{i} \rightarrow \mathrm{c}}$ rate of 11\%. This results in a $\Delta^{\text{Acc}}$ of 9.2\%, the highest among all baselines.

\begin{table}[t]
\centering
\resizebox{\columnwidth}{!}{%
\begin{tabular}{lccc}
\toprule
\textbf{Method} & \textbf{Accuracy@t1} & \textbf{Accuracy@t2} & \textbf{$\Delta^\textbf{Acc}$(t1, t2)} \\ \midrule
\multicolumn{4}{c}{\textbf{MBPP}} \\ \midrule
Boost model & 10.4 / 16.6 & 11.0 / 20.2 & 0.6 / 3.6 \\
Turn-SFT & 23.2 / 45.0 & 23.6 / 46.0 & 0.4 / 1.0 \\
Self-Corrector & 27.6 / 46.4 & 25.8 / 47.0 & $-$1.8 / 0.6 \\
ReVISE & 29.0 / \textbf{48.6} & 30.8 / 47.6 & 0.8 / $-$1.2 \\ 
Turn-RL & 16.6 / 36.4 & 16.4 / 55.4 & $-$0.2 / 19.0 \\\hdashline
\textsc{CoCoS} & \textbf{57.2} / 48.2 & \textbf{59.4} / \textbf{60.4} & 2.2 / 12.2 \\
\midrule
\multicolumn{4}{c}{\textbf{HumanEval}} \\ \midrule
Boost model & 15.2 / 21.3 & 15.2 / 21.3 & 0.0 / 0.0 \\
Turn-SFT & 14.6 / 19.5 & 12.8 / 22.0 & $-$1.9 / 2.5 \\
Self-Corrector & 11.0 / 20.1 & 12.8 / 22.6 & 1.9 / 2.4 \\
ReVISE & 13.4 / \textbf{23.8} & 11.9 / 19.0 & $-$1.9 / $-$4.3 \\ 
Turn-RL & 8.5 / 23.2 & 8.5 / \textbf{25.6} & 0.0 / 2.4 \\\hdashline
\textsc{CoCoS} & \textbf{34.1} / 22.6 & \textbf{39.6} / 25.0 & 5.5 / 2.4 \\ \midrule
\multicolumn{4}{c}{\textbf{ODEX}} \\ \midrule
Boost model & 13.2 / 27.3 & 12.8 / 27.6 & $-$0.5 / $-$0.3 \\
Turn-SFT & 15.7 / \textbf{29.8} & 15.3 / 28.7 & $-$0.4 / $-$1.1 \\
Self-Corrector & 13.2 / 27.3 & 14.4 / 28.0 & 1.1 / 0.7 \\
ReVISE & 9.6 / 21.2 & 11.4 / 22.8 & 1.8 / 1.6 \\
Turn-RL & 19.8 / 23.6 & 21.9 / 30.5 & 2.1 / 6.9 \\\hdashline
\textsc{CoCoS} & \textbf{23.2} / 26.2 & \textbf{25.1} / \textbf{31.4} & 1.8 / 5.2 \\
\bottomrule
\end{tabular}%
}
\caption{Accuracy comparison across different models. Results are reported as Llama-3.2-1B / deepseek-coder-1.3B. Percentage units (\%) are omitted from the table. The full results are provided in Appendix~\ref{tab:detailed result}.}
\label{tab:main_result_other}
\end{table}
\subsection{Evaluation in broader scenarios}
To evaluate the generalization of our approach, we report additional results on alternative models and unseen datasets in Table~\ref{tab:main_result_other}.
\textsc{CoCoS} outperforms other baselines even when using SLMs other than Qwen as the backbone. For instance, using Llama as the backbone, it achieves the highest correction rate on MBPP, with a $\Delta^\text{Acc}$ of 12.2\%.
Additionally, The SFT-based baselines (Turn-SFT, Self-Corrector, and ReVISE) often fail to generalize beyond their training distribution~\citep{chu2025sft}, as evidenced by the comparable performance to their backbones.
On the MBPP, SFT-based baselines show up to a 2× improvement over the backbone in Accuracy@t2. However, in unseen settings, their correction performance is comparable to that of the Boost model or even degrades. In contrast, \textsc{CoCoS} consistently yields positive values on the $\Delta^\text{Acc}$ metric, demonstrating stable improvement across turns.
\begin{figure}[t]
\centering
\includegraphics[width=\columnwidth,keepaspectratio]{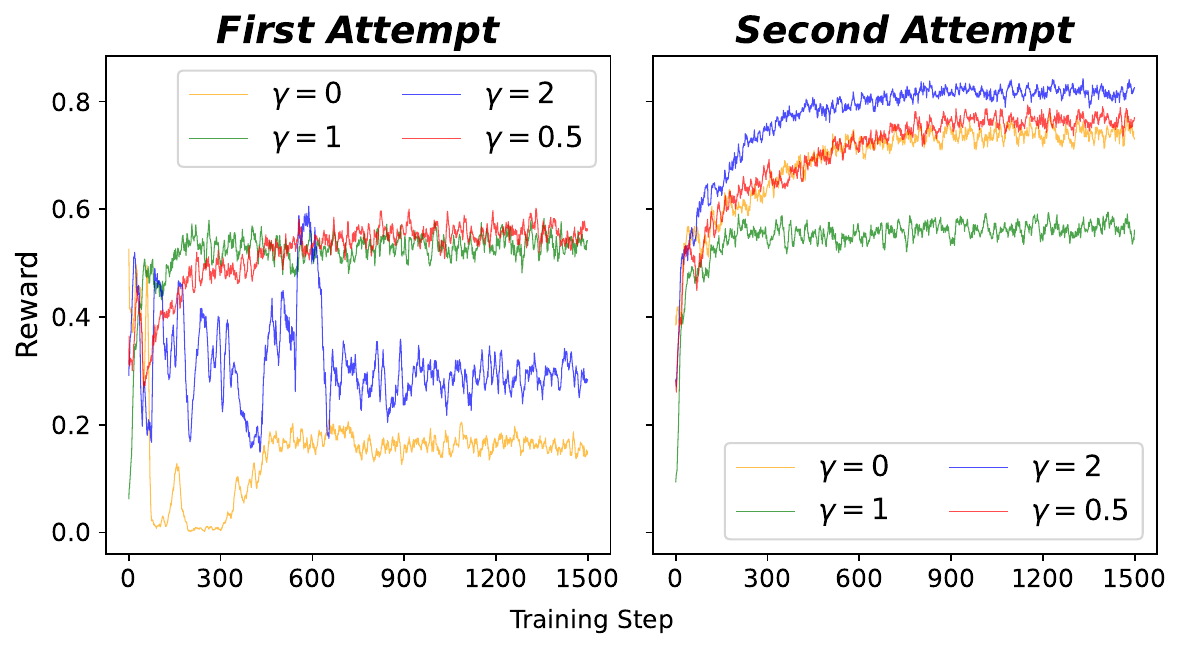}
\caption{Learning curves under varying values of discount factor $\gamma$.}
\label{fig:discount_factor}
\end{figure}

\section{Analyses}
In this section, we analyze the discount factor in the accumulated reward function (\S\ref{sec:discount factor}), compare progressive and binary reward schemes (\S\ref{sec:progressive reward}), examine multi-turn correction beyond two turns (\S\ref{sec:turn-wise}), evaluate the model’s robustness to varied correction instructions during inference (\S\ref{sec:auxiliary instruction}), and conduct case studies on representative examples (\S\ref{sec:case study}).

\begin{figure*}[t]
\centering
\includegraphics[width=\textwidth,keepaspectratio]{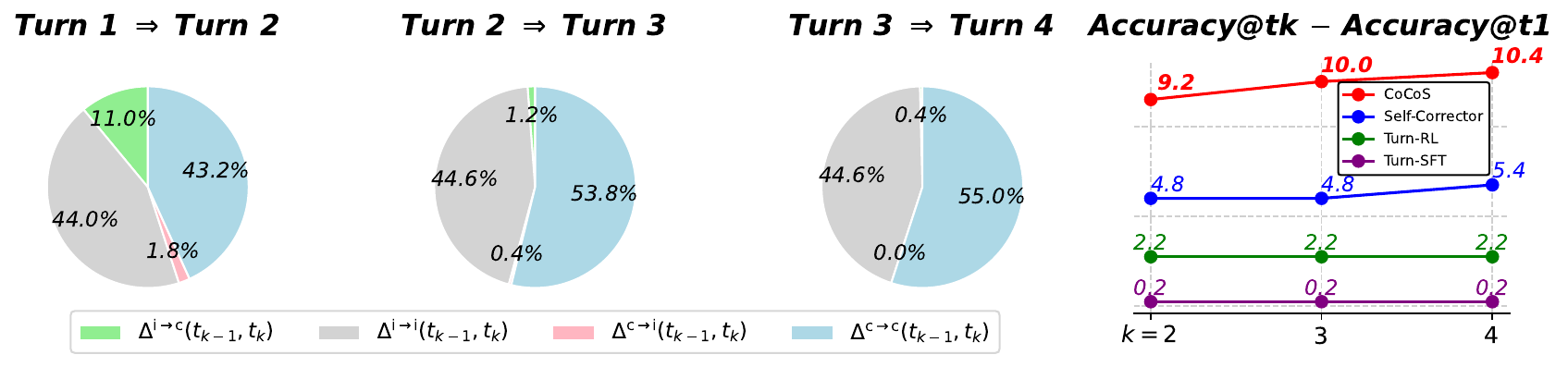}
\caption{Turn-wise metric changes on MBPP using the Qwen model. The pie charts represent the distribution of transition types: $\Delta^{\mathrm{i} \rightarrow \mathrm{c}}$, $\Delta^{\mathrm{i} \rightarrow \mathrm{i}}$, $\Delta^{\mathrm{c} \rightarrow \mathrm{i}}$, and $\Delta^{\mathrm{c} \rightarrow \mathrm{c}}$ across successive turns. The line plot on the right shows the Accuracy@tk $-$ Accuracy@t1 as the number of turns increases.}
\label{fig:pie_chart}
\end{figure*}

\subsection{Discount factor}\label{sec:discount factor}

As shown in Equation~\ref{eq:reward function}, when $T=2$, the reward function is formulated as $R(\hat{y}_{1:2})=(r_2 - r_1) + \gamma \cdot r_1$. We investigate the effect of varying the discount factor $\gamma$ on training behavior, with learning curves shown in Figure~\ref{fig:discount_factor}. When $\gamma=0$, training focuses on maximizing the reward difference $r_2 - r_1$, often by deliberately decreasing $r_1$. When $\gamma=1$, the reward depends only on the second response, disregarding the first turn. As a result, responses across turns become tightly coupled, leading to poor coverage in subsequent iterations~\citep{kumar2025training}. Under this setting, the model fails to develop the ability to self-correct.

Additionally, setting $\gamma = 2$ to jointly optimize both responses causes instability during training. In this case, it becomes unclear which of the two responses contributed positively to the total reward. This situation is called the credit assignment problem in MDPs~\citep{pignatelli2024a}. This is beyond the scope of the current research and is not further addressed in this work. Consequently, we set $\gamma=0.5$ to balance the quality of the initial response and the refinement achieved in the second.

\subsection{Progressive reward}\label{sec:progressive reward}
\begin{table}[t]
\centering
\resizebox{\columnwidth}{!}{%
\begin{tabular}{lcccc}
\toprule
\textbf{Dataset} & \textbf{Accuracy@t1} & \textbf{Accuracy@t2} & \textbf{$\Delta_{\text{unit}}^{\mathrm{i} \rightarrow \mathrm{c}}$} & \textbf{$\Delta_{\text{unit}}^{\mathrm{c} \rightarrow \mathrm{i}} \downarrow$} \\ \midrule
\multicolumn{5}{c}{\textbf{Binary / Progressive}} \\ \midrule
MBPP & 45.0 / 45.0 & 51.2 / \textbf{54.2} & 10.5 / \textbf{11.9} & 3.4 / \textbf{1.9} \\
HumanEval & 24.4 / \textbf{29.3} & 27.4 / \textbf{32.3} & \textbf{9.6} / 4.7 & 5.0 / \textbf{3.1} \\
ODEX & 23.0 / 23.0 & 25.7 / \textbf{28.9} & 9.8 / \textbf{34.2} & 3.9 / \textbf{0.0} \\
\bottomrule
\end{tabular}%
}
\caption{Comparison of progressive vs. binary reward schemes. Qwen2.5-1.5B models are trained under each reward setting and evaluated on MBPP.}
\label{tab:fine-grained}
\end{table}
We compare the progressive reward (as defined in Equation~\ref{eq:progressive reward}) with a binary assessment scheme. To enable fine-grained evaluation, we assess the result of each unit test case individually. Specifically, \textbf{$\Delta_{\text{unit}}^{\mathrm{c} \rightarrow \mathrm{i}}$} measures how many test cases drop from passing to failing between turns. Under the progressive reward, a transition from 2 passes to 1 is penalized for discouraging such behavior, whereas the binary reward assigns 0 in both cases, failing to capture the distinction.

Table~\ref{tab:fine-grained} presents the results of models trained with the two reward schemes. The progressive reward demonstrates the strong preservation of correct unit test cases. For instance, on the ODEX, 34\% of test cases show improvement, with a 100\% preservation rate. In comparison, the binary reward causes around 40\% of previously passed test cases to fail in subsequent turns. These results highlight the importance of fine-grained assessment, which has been overlooked in the multi-turn scenario.

\subsection{Turn-wise changes}\label{sec:turn-wise}
We analyze how \textsc{CoCoS} affects the distribution of correct and incorrect responses over multiple turns. Figure~\ref{fig:pie_chart} presents the distribution and trend of changes in response correctness. First, the pie charts show that the proportion of $\Delta^{\mathrm{c} \rightarrow \mathrm{c}}$ gradually increases with each turn. This indicates that once a correct response is generated, the model maintains increasing consistency in subsequent turns.

Second, the proportion of $\Delta^{\mathrm{c} \rightarrow \mathrm{i}}$ remains consistently small, and it continues to decrease over successive turns. This suggests that \textsc{CoCoS} enables a model to refine its code iteratively, maintaining confidence in correct responses while effectively reducing errors over time. Finally, the line plot illustrates the cumulative net gain in $\Delta^\text{Acc}$. \textsc{CoCoS} achieves steady gains over the baseline across turns, even though it starts from a high Accuracy@t1.

\begin{table}[t]
\small
\centering
\resizebox{\columnwidth}{!}{%
\begin{tabular}{lcc}
\toprule
\textbf{Model} & \textbf{Accuracy@t2} & \textbf{$\Delta^\textbf{Acc}$(t1, t2)} \\ \midrule
\multicolumn{3}{c}{\textbf{Template / Fixed}} \\ \midrule
Qwen2.5-1.5B & 52.8 / 54.2 & 7.2 / 9.2 \\
Llama-3.2-1B & 58.6 / 59.4 & 1.4 / 2.2 \\
deepseek-coder-1.3B & 60.2 / 60.4 & 12.0 / 12.2 \\
\bottomrule
\end{tabular}%
}
\caption{Evaluation results with varied instruction rephrasings on the MBPP dataset.}
\label{tab:auxiliary instruction}
\end{table}

\subsection{Auxiliary instruction}\label{sec:auxiliary instruction}
To evaluate the generalization ability of our method across different prompts, we intentionally use a different instruction prompt at test time than the one used during training (as defined in Appendix~\ref{prompt:main}). Instead of the fixed training prompt, we manually rephrase it into five alternative versions for the correction task.
The results are shown in Table~\ref{tab:auxiliary instruction}, and the rephrased instructions are provided in Appendix~\ref{prompt:auxiliary instruction}. While the fixed prompt used during training yields slightly better overall performance, the model exhibits no substantial degradation under instruction variation. The average performance drop at the second attempt is only 0.8\%, indicating robustness to diverse prompt variations.

\subsection{Case Study}\label{sec:case study}
To analyze the types of mistakes that \textsc{CoCoS} makes and how it corrects them, we conduct three case studies. The observed error types include: (1) naming errors, (2) complex logic errors, and (3) incorrect usage of code libraries. In the main text, we focus on the first category, while detailed discussions of the latter two are provided in the Appendix~\ref{apdx:case_study}.

\begin{lstlisting}[language=Python,linewidth=\columnwidth]
# Test case
assert test_duplicate(([1,2,3,4,5]))==False
assert test_duplicate(([1,2,3,4, 4]))==True
assert test_duplicate([1,1,2,2,3,3,4,4,5])==True

# Initial response
# wrong function name; won't match the test case
def contains_duplicate(nums):
    return len(nums)!= len(set(nums))
\end{lstlisting}
\vspace{-1mm}
\begin{lstlisting}[language=Python,backgroundcolor=\color{ll1},caption=Example of correcting a misnamed function,label=code:naming,linewidth=\columnwidth]
# Corrected response
def test_duplicate(arr):
    return len(arr)!= len(set(arr))
\end{lstlisting}
Code~\ref{code:naming} highlights an issue related to function name generation. Given the test cases, the model is expected to generate a function name that allows the code to execute correctly. However, in some instances, the model fails to produce the appropriate function name. In such cases, even if the core logic is correctly implemented, the test cases cannot be executed, and the result is considered a generation failure. In the following turn, when prompted to correct the previous error, the model is able to preserve the correct logic and revise only the function name, thereby resolving the issue.
\section{Conclusion}
In this paper, we investigated self-correction methods for enabling SLMs to revise their own generated code. Our findings show that, despite their effectiveness in proprietary models, existing prompting- and training-based approaches fall short when applied to SLMs. To address these shortcomings, we propose \textsc{CoCoS}, which introduces an RL reward scheme tailored to the multi-turn code generation setting. By incorporating fine-grained assessment and an accumulated reward function, \textsc{CoCoS} demonstrates both successful self-correction and improved initial response quality through trajectory-level optimization across diverse SLMs and previously unseen scenarios.
\nocite{*}
\section*{Acknowledgments}
This work was supported by the IITP(Institute of Information \& Coummunications Technology Planning \& Evaluation)-ITRC(Information Technology Research Center) grant funded by the Korea government(Ministry of Science and ICT)(IITP-2025-RS-2024-00437866, 47.5\%). This work was supported by Smart HealthCare Program funded by the Korean National Police Agency(KNPA) (No. RS-2022-PT000186, 47.5\%). This work was supported by Institute of Information \& communications Technology Planning \& Evaluation (IITP) grant funded by the Korea government(MSIT) (No.RS-2019-II191906, Artificial Intelligence Graduate School Program(POSTECH), 5\%).

\section*{Limitations}
Online reinforcement learning incurs a cost at every action step, as the model learns by interacting with the environment in real-time. Due to infrastructural constraints, we limited \textsc{CoCoS} training to only two turns. In the turn-wise analysis section, \textsc{CoCoS} consistently demonstrated accuracy gains as the number of turns increased. These results suggest that increasing the number of training turns has the potential to yield further improvements. Therefore, future work should extend our approach to support multi-turn training beyond two turns, potentially incorporating more cost-efficient strategies such as offline reinforcement learning. 

For the same reason, our experiments were conducted only on a 1B-scale small language model. While we demonstrated that our methodologies are effective on small models, our approach is not limited to this setting. Moreover, evaluating its scalability to models larger than 1B parameters remains an interesting aspect to observe in future work.

\section*{Ethical Considerations}
In our research, we use datasets such as KodCode~\citep{xu2025kodcode}, MBPP~\citep{austin2021program}, HumanEval~\citep{chen2021evaluating}, and ODEX~\citep{wang-etal-2023-execution}, which are licensed under CC BY-NC 4.0, CC BY 4.0, MIT, and CC BY-SA 4.0, respectively. The models GPT-4.1~\citep{gpt4.1}, Qwen2.5-1.5B~\citep{yang2024qwen2}, Llama-3.2-1B~\citep{grattafiori2024llama}, and deepseek-coder-1.3B~\citep{guo2024deepseek} are licensed under OpenAI, Apache-2.0, Llama 3.2 Community, and deepseek-license, respectively. All models were used strictly for research purposes, and no artifacts were utilized beyond the scope of the study.
\bibliography{custom}

\clearpage
\appendix
\onecolumn
\section{Additional Experiment on SETS}\label{apdx:sets}
\begin{table}[h]
\centering
\begin{tabular}{l||cccc}
\toprule
\textbf{Verifier} & \textbf{True Positive} & \textbf{False Positive} & \textbf{True Negative} & \textbf{False Negative} \\\midrule
Qwen2.5-1.5B & 35.1\% & 54.1\% & 7.2\% & 3.6\% \\
GPT-4.1 & 27.6\% & 22.2\% &43.1\% & 7.1\% \\\bottomrule
\end{tabular}
\caption{Verification accuracy based on different verifiers. \textbf{\textit{The Qwen verifier shows near-random performance, while the teacher verifier reaches 70\% accuracy.}}}
\label{tab:verifier}
\end{table}
To assess the accuracy of the verifier module, we directly compare the verification results from SETS with the test case pass rate. Specifically, we use the test cases as the ground-truth labels and the verifier as the predictor and report the confusion matrix in Table~\ref{tab:verifier}. The Qwen verifier achieves an accuracy of 42\%, which is close to random performance in distinguishing correct and incorrect responses. This indicates that small-scale Qwen fails to provide effective feedback, thereby harming rather than improving the initial response. In contrast, the teacher verifier reaches a higher verification accuracy of 70\%. However, despite employing a costly verifier with relatively high accuracy, the results in Table~\ref{tab:prompting-based} show only a marginal 5\% gain over simple prompting.
\section{Baseline Details}\label{apdx:baseline}
In this section, we provide a detailed overview of the baseline used in our experimental evaluations. To implement the baseline, we constructed the training datasets, and to illustrate this process, we provide the 1-turn and 2-turn data samples. The examples are as follows:
\begin{prompt}[title={Data Sample \thetcbcounter: 1-Turn Example}, label=prompt:1-turn]
You are an expert Python programmer, and here is your task: {{problem}} Your code should pass these tests:\\

\{\{test cases\}\}\\

[BEGIN]

\textcolor{blue}{\{\{correct code\}\}}

[DONE]
\end{prompt}
\begin{prompt}[title={Data Sample \thetcbcounter: 2-Turn Example}, label=prompt:2-turn]
You are an expert Python programmer, and here is your task: {{problem}} Your code should pass these tests:\\

{{test cases}}\\

[BEGIN]

\textcolor{red}{\{\{incorrect code\}\}}

[DONE]\\

There might be an error in the code above because of a lack of understanding of the question.

Please correct the error, if any, and rewrite the solution. Only output the final correct Python program!\\

[CORRECT]

\textcolor{blue}{\{\{correct code\}\}}

[DONE]
\end{prompt}

\paragraph{Turn-SFT.}
We generate training data using pre-trained models according to the format specified in \S\ref{prompt:main}. For each problem, we sample 10 candidate code solutions and determine their correctness based on the results of corresponding unit test cases. We then organize this data into two separate datasets: 1-turn and 2-turn. We combine the 1-turn and 2-turn datasets in a 1:1 ratio to match the data scale of other baseline methods, allowing the model to optimize initial code generation and correction capabilities.

\paragraph{Self-Corrector~\citep{welleck2023generating}.}
Self-Corrector acts as a specialized plug-in designed exclusively for the correction task. It trains as a separate component from the initial response model and generates only corrected outputs. For training data, we only use 2-turn data, which removes the initial code generation phase and focuses entirely on correction. We use the same LLM for both the response and correction LLM, but we freeze the response LLM during training while updating the correction LLM.

\paragraph{ReVISE~\citep{lee2025revise}.}
ReVISE uses a dedicated \texttt{[refine]} token to trigger the correction phase explicitly. The model generates either a \texttt{[refine]} token or an \texttt{[eos]} token. When the \texttt{[refine]} token is produced, the model enters the correction phase and continues generating outputs until it produces the \texttt{[eos]} token, which signals the end of the correction process. This design allows the model to complete self-correction in a single pass without intermediate user inputs, which eliminates the need for multiple generation stages. We generate data through sampling-based decoding and train the model by applying SFT loss and Direct Preference Optimization~\citep{rafailov2023direct} loss during the backward pass.

\paragraph{Turn-RL.}
Turn-RL is trained to generate the \texttt{{{correct code}}} following the \texttt{[CORRECT]} token in 2-turn data, while the 1-turn data is pre-sampled and remains fixed during training. The training objective and reward function are shared with \textsc{CoCoS}.
\section{Experimental Details}\label{apdx:detail}
\subsection{Policy Optimization}
We adopt the policy gradient method using the REINFORCE Leave-One-Out (RLOO) estimator~\citep{ahmadian-etal-2024-back}. RLOO provides a simple and efficient baseline, and can further reduce variance when multiple online samples are available by using each sample’s reward as a baseline for others and averaging gradient estimates. We extend this approach to the multi-turn setting to train \textsc{CoCoS}.
Our training objective, as defined in Equation~\ref{eq:cocos}, is as follows:
\begin{equation}\nonumber
    J(\theta) = \mathbb{E}_{(x,y) \sim \mathcal{D},\; \hat{y}\ \sim \pi_\theta} \Big[R(\hat{y}_{1:T}) - \beta \sum_{t=1}^{T}\mathbb{D}_{KL}\big(\pi_\theta(\cdot \mid s_t) \;\|\; \pi_{ref}(\cdot \mid s_t)\big)\Big]
\end{equation}
We then optimize the expected return using policy gradients with the RLOO estimator:
\begin{equation}\nonumber
    \nabla_\theta J(\theta) = \mathbb{E}_{(x,y) \sim \mathcal{D},\; \hat{y}\ \sim \pi_\theta} \big[R'(\hat{y}_{1:T}) \sum_{t=1}^{T}\nabla_\theta \log \pi_\theta(\hat{y}_t \mid s_t)\big] - \beta \sum_{t=1}^{T} \nabla_\theta \mathbb{D}_{KL}\big(\pi_\theta(\cdot \mid s_t) \;\|\; \pi_{ref}(\cdot \mid s_t)\big)
\end{equation}
Here, $R'(\hat{y}_{1:T})$ denotes the leave-one-out baseline-adjusted reward computed over the entire trajectory. For a sampled trajectory $\hat{y}_{1:T}^{(i)}$, the trajectory-level reward is defined as:
\begin{equation}\nonumber
R(\hat{y}_{1:T}^{(i)}) = \gamma^{T-1} r_1^{(i)} + \sum_{t=2}^T \gamma^{T-t} (r_t^{(i)} - r_{t-1}^{(i)})
\end{equation}
where $r_t^{(i)}$ is the scalar reward associated with the model’s response $\hat{y}_t^{(i)}$ at turn $t$. Given $k$ such sampled trajectories, we define the leave-one-out adjusted reward as:
\begin{equation}\nonumber
    R'(\hat{y}_{1:T}^{(i)}) = R(\hat{y}_{1:T}^{(i)}) - \frac{1}{k-1} \sum_{j \ne i} R(\hat{y}_{1:T}^{(j)})
\end{equation}
This estimator leverages the rewards of the other $k-1$ samples as a baseline, reducing variance while maintaining unbiasedness. Accordingly, the full policy gradient is computed as:
\begin{equation}\nonumber
\begin{split}
    \therefore \nabla_\theta J(\theta) = \mathbb{E}_{(x,y) \sim \mathcal{D}} \bigg[ \frac{1}{k} \sum_{i=1}^{k} \Big(R'(\hat{y}_{1:T}^{(i)}) & \sum_{t=1}^{T}\nabla_\theta \log \pi_\theta(\hat{y}_t^{(i)} \mid s_t^{(i)}) \\
    & - \beta \sum_{t=1}^{T} \nabla_\theta \mathbb{D}_{KL}\big(\pi_\theta(\cdot \mid s_t^{(i)}) \;\|\; \pi_{ref}(\cdot \mid s_t^{(i)})\big)\Big)\bigg]
\end{split}
\end{equation}

\subsection{Hyperparameter}
We trained the model via the Transforming Reinforcement Learning~\citep{10546317}. For both baselines and \textsc{CoCoS} training, we used 4 NVIDIA A100-SXM4-80GB. For inference, all experiments were conducted on a single NVIDIA RTX 6000 Ada Generation. The hyperparameters used in our experiments are listed in Table~\ref{tab:hyperparameters}. Rather than conducting an extensive hyperparameter search, we trained all models with a unified set of hyperparameters for \textsc{CoCoS}, regardless of the backbone model.
\begin{table}[h]
\centering
\begin{tabular}{l|cc}
\toprule
\textbf{Hyperparameters} & \textbf{Boost model} & \textbf{\textsc{CoCoS}} \\ \midrule
Dataset & KodCode & MBPP \\
Global batch size & 256 & 128 \\
Optmizer & AdamW & Adam \\
Weight decay & 0.1 & - \\
Learning rate & 2e-5 & 1e-5 \\
LR scheduler & cosine & cosine \\
Training steps & 1000 & 1500 \\
Sampling temperature & - & 0.9 \\
KL coefficient ($\beta$) & - & 0.01 \\
RLOO samples ($k$) & - & 2 \\
\bottomrule
\end{tabular}
\caption{Hyperparameters used in our experiments.}
\label{tab:hyperparameters}
\end{table}

\subsection{Dataset}
\begin{table}[h]
\centering
\begin{tabular}{cc|cccc|c|c}
\toprule
\multicolumn{2}{c|}{\textbf{KodCode}} & \multicolumn{4}{c|}{\textbf{MBPP}} & \textbf{HumanEval} & \textbf{ODEX}\\
Train & Validation & Train & Validation & Test & Few-shot & Test & Test\\ 
\midrule
144,068 & 36,018 & 374 & 90 & 500 & 10 & 164 & 439\\
\bottomrule
\end{tabular}
\caption{Dataset sizes used in our experiments.}
\label{tab:dataset}
\end{table}
We report all the datasets used in our experiments in Table~\ref{tab:dataset}. The KodCode~\citep{xu2025kodcode} dataset was used for training the boost model, and we preprocessed the data to extract only samples that can be converted into our 2-turn format. As a result, out of 484k samples, only 180k were used for training. These samples were then randomly split into an 8:2 train/validation set. In MBPP~\citep{austin2021program}, given that our boost models were already pre-trained, we excluded few-shot data. Instead of using separate validation data, we selected the final model checkpoint based on the highest reward achieved during training.
\clearpage
\section{Detailed Results}\label{tab:detailed result}
\subsection{MBPP}
\begin{table}[h]
\centering
\begin{tabular}{lccccc}
\toprule
\textbf{Method} & \textbf{Accuracy@t1} & \textbf{Accuracy@t2} & $\Delta^{\mathrm{i} \rightarrow \mathrm{c}}$\textbf{(t1, t2)} & $\Delta^{\mathrm{c} \rightarrow \mathrm{i}}$\textbf{(t1, t2)}$\downarrow$ & \textbf{$\Delta^\textbf{Acc}$(t1, t2)}\\
\midrule
\multicolumn{6}{c}{\textbf{Qwen2.5-1.5B}} \\ \midrule
Boost model & 9.0\% & 10.2\% & 1.2\% & 0.0\% & 1.2\% \\
Turn-SFT & 44.2\% & 44.4\%  & 1.4\% & 1.2\% & 0.2\% \\
Self-Corrector & 43.8\% & 48.8\%  & 11.2\% & 6.2\% & 5.0\%\\
ReVISE & 34.6\% & 41.2\% & 16.1\%  & 4.9\% & 6.4\% \\
Turn-RL & 46.4\% & 48.6\% & 4.0\% & 1.8\% & 2.2\% \\\hdashline
\textsc{CoCoS} & 45.0\% & 54.2\%  & 11.0\% & 1.8\% & 9.2\% \\
\midrule
\multicolumn{6}{c}{\textbf{Llama-3.2-1B}} \\\midrule
Boost model & 10.4\% & 11.0\%  & 0.6\% & 0.0\% & 0.6\% \\
Turn-SFT & 23.2\% & 23.6\%  & 1.2\% & 0.8\% & 0.4\% \\
Self-Corrector & 27.6\% & 25.8\%  & 5.8\% & 7.6\% & $-$1.8\% \\
ReVISE & 29.0\% & 30.8\%  & 5.8\% & 5.0\% & 0.8\% \\
Turn-RL & 16.6\% & 16.4\% & 0.2\% & 0.4\% & $-$0.2\% \\\hdashline
\textsc{CoCoS} & 57.2\% & 59.4\%  & 3.6\% & 1.4\% & 2.2\% \\
\midrule
\multicolumn{6}{c}{\textbf{deepseek-coder-1.3B}} \\\midrule
Boost model & 16.6\% & 20.2\%  & 3.6\% & 0.0\% & 3.6\% \\
Turn-SFT & 45.0\% & 46.0\%  & 4.0\% & 3.0\% & 1.0\% \\
Self-Corrector & 46.4\% & 47.0\%  & 9.4\% & 8.8\% & 0.6\% \\
ReVISE & 48.6\% & 47.6\%  & 3.2\% & 4.4\% & $-$1.2\% \\
Turn-RL & 36.4\% & 55.4\% & 19.8\% & 0.8\% & 19.0\%  \\\hdashline
\textsc{CoCoS} & 48.2\% & 60.4\%  & 13.0\% & 0.8\% & 12.2\% \\
\bottomrule
\end{tabular}
\caption{Results on the MBPP dataset.}
\label{tab:mbpp}
\end{table}
\clearpage
\subsection{HumanEval}
\begin{table}[h]
\centering
\begin{tabular}{lccccc}
\toprule
\textbf{Method} & \textbf{Accuracy@t1} & \textbf{Accuracy@t2}  & $\Delta^{\mathrm{i} \rightarrow \mathrm{c}}$\textbf{(t1, t2)} & $\Delta^{\mathrm{c} \rightarrow \mathrm{i}}$\textbf{(t1, t2)}$\downarrow$ & \textbf{$\Delta^\textbf{Acc}$(t1, t2)} \\
\midrule
\multicolumn{6}{c}{\textbf{Qwen2.5-1.5B}} \\\midrule
Boost model & 24.4\% & 25.5\%  & 1.1\% & 0.0\% & 1.1\% \\
Turn-SFT & 18.9\% & 20.1\%  & 5.5\% & 4.3\% & 1.2\% \\
Self-Corrector & 23.2\% & 25.6\%  & 6.1\% & 3.7\% & 2.4\% \\
ReVISE & 18.3\% & 15.9\%  & 3.2\% & 4.5\% & $-$1.3\% \\
Turn-RL & 27.4\% & 28.7\% & 1.8\% & 0.6\% & 1.2\% \\\hdashline
\textsc{CoCoS} & 29.3\% & 32.3\%  & 3.0\% & 0.0\% & 3.0\% \\
\midrule
\multicolumn{6}{c}{\textbf{Llama-3.2-1B}} \\\midrule
Boost model  & 15.2\% & 15.2\%  & 0.6\% & 0.6\% & 0.0\% \\
Turn-SFT & 14.6\% & 12.8\%  & 1.8\% & 3.7\% & $-$1.9\% \\
Self-Corrector & 11.0\% & 12.8\%  & 4.9\% & 3.0\% & 1.9\% \\
ReVISE & 13.4\% & 11.9\%  & 3.7\% & 5.6\% & $-$1.9\% \\
Turn-RL & 8.5\% & 8.5\% & 0.0\% & 0.0\% & 0.0\% \\\hdashline
\textsc{CoCoS} & 34.1\% & 39.6\%  & 6.1\% & 0.6\% & 5.5\% \\
\midrule
\multicolumn{6}{c}{\textbf{deepseek-coder-1.3B}} \\\midrule
Boost model & 21.3\% & 21.3\%  & 0.0\% & 0.0\% & 0.0\% \\
Turn-SFT & 19.5\% & 22.0\%  & 5.5\% & 3.0\% &2.5\% \\
Self-Corrector & 20.1\% & 22.6\%  & 7.9\% & 5.5\% & 2.4\% \\
ReVISE & 23.8\% & 19.0\%  & 3.1\% & 7.4\% & $-$4.3\% \\
Turn-RL & 23.2\% & 25.6\% & 3.0\% & 0.6\% & 2.4\% \\\hdashline
\textsc{CoCoS} & 22.6\% & 25.0\% & 3.0\% & 0.6\% & 2.4\% \\ \bottomrule
\end{tabular}
\caption{Results on the HumanEval dataset.}
\label{tab:humaneval}
\end{table}
\clearpage
\subsection{ODEX}
\begin{table}[h]
\centering
\begin{tabular}{lccccc}
\toprule
\textbf{Method} & \textbf{Accuracy@t1} & \textbf{Accuracy@t2}  & $\Delta^{\mathrm{i} \rightarrow \mathrm{c}}$\textbf{(t1, t2)} & $\Delta^{\mathrm{c} \rightarrow \mathrm{i}}$\textbf{(t1, t2)}$\downarrow$ & \textbf{$\Delta^\textbf{Acc}$(t1, t2)} \\
\midrule
\multicolumn{6}{c}{\textbf{Qwen2.5-1.5B}} \\\midrule
Boost model & 21.2\% & 20.5\%  & 0.0\% & 0.7\% & $-$0.7\% \\
Turn-SFT & 25.5\% & 25.7\%  & 0.5\% & 0.2\% & 0.3\% \\
Self-Corrector & 21.2\% & 27.1\%  & 7.5\% & 1.6\% & 5.9\% \\
ReVISE & 18.5\% & 19.0\% & 8.7\%  & 6.0\% & 0.5\% \\ 
Turn-RL & 10.9\% & 24.6\% & 14.1\% & 0.5\% & 13.6\% \\\hdashline
\textsc{CoCoS} & 23.0\% & 28.9\%  & 5.9\% & 0.0\% & 5.9\% \\
\midrule
\multicolumn{6}{c}{\textbf{Llama-3.2-1B}} \\\midrule
Boost model & 13.2\% & 12.8\%  & 0.2\% & 0.7\% & $-$0.5\% \\
Turn-SFT & 15.7\% & 15.3\%  & 0.7\% & 1.1\% & $-$0.4\% \\
Self-Corrector & 13.2\% & 14.4\%  & 5.9\% & 4.8\% & 1.1\% \\
ReVISE & 9.6\% & 11.4\%  & 4.8\% & 3.0\% & 1.8\% \\
Turn-RL & 19.8\% & 21.9\% & 2.1\% & 0.0\% & 2.1\% \\\hdashline
\textsc{CoCoS} & 23.2\% & 25.1\%  & 2.3\% & 0.5\% & 1.8\% \\
\midrule
\multicolumn{6}{c}{\textbf{deepseek-coder-1.3B}} \\\midrule
Boost model & 27.3\% & 27.6\%  & 0.5\% & 0.2\% & $-$0.3\% \\
Turn-SFT & 29.8\% & 28.7\%  & 0.0\% & 1.1\% & $-$1.1\% \\
Self-Corrector & 27.3\% & 28.0\%  & 3.9\% & 3.2\% & 0.7\% \\
ReVISE & 21.2\% & 22.8\% & 8.1\%  & 4.8\% & 1.6\% \\
Turn-RL & 23.6\% & 30.5\% & 6.9\% & 0.0\% & 6.9\% \\\hdashline
\textsc{CoCoS} & 26.2\% & 31.4\%  & 5.9\% & 0.7\% & 5.2\% \\
\bottomrule
\end{tabular}
\caption{Results on the ODEX dataset.}
\label{tab:odex}
\end{table}
\clearpage
\section{Additional SCoRe Experiment on Another Model}\label{apdx:score}

\begin{figure}[h]
\centering
\includegraphics[width=.6\textwidth,keepaspectratio]{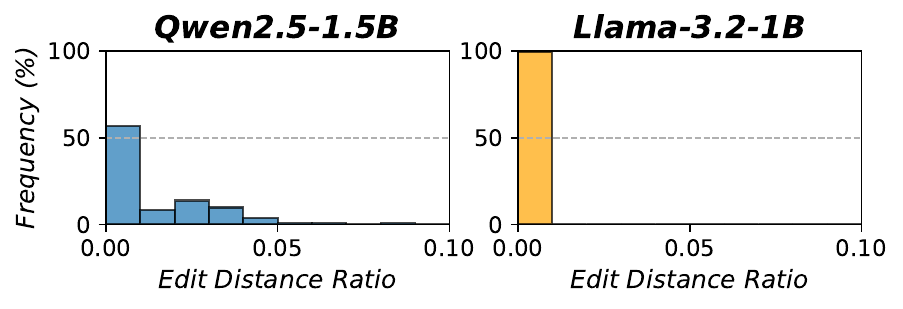}
\caption{
Edit distance ratio distributions between first and second responses for Qwen2.5-1.5B and Llama-3.2-1B in the 3-shot setting.
}
\label{fig:edit distance appendix}
\end{figure}

\begin{figure}[h]
\centering
\includegraphics[width=\textwidth,keepaspectratio]{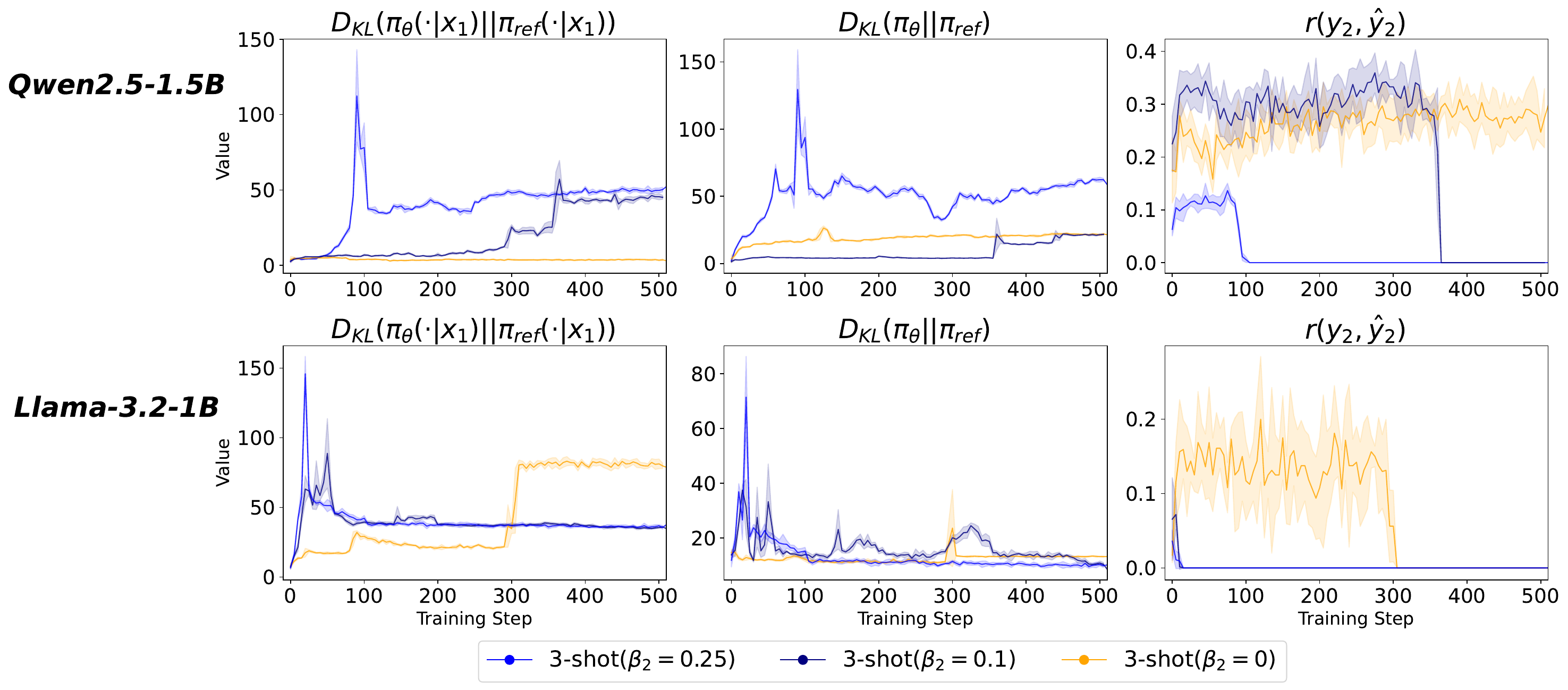}
\caption{Learning curves for SCoRe, including additional results for the small-scale Llama model.}
\label{fig:score appendix}
\end{figure}

We further analyze the training trajectory of SCoRe using a small-scale Llama model in addition to the Qwen model. Prior to this, we measured the edit distance ratio. Following the same procedure as in \S\ref{sec:score}, we compute the edit distance ratio for samples where the second response is incorrect. The results are shown in Figure~\ref{fig:edit distance appendix}. Remarkably, the Llama model exhibits an extreme tendency to make no changes to its responses, with a no-edit rate of 99\%. We then compare the SCoRe training trajectories of Qwen and Llama in Figure~\ref{fig:score appendix}. As Qwen has already been described in \S\ref{sec:score}, we omit further discussion here. In the case of Llama, applying KL-regularization leads to a rapid convergence of the reward to zero, similar to what was observed with Qwen. Even without KL regularization, the Llama model eventually collapses. We speculate that this shows a limitation of training SLMs with binary rewards. When the policy is trained with binary rewards, the model receives only a pass or fail signal for $r(y_2, \hat{y}_2)$, which severely limits the space the policy can explore during training. As a result, in the case of SLMs that receive persistently low rewards, the policy ultimately collapses, failing to explore diverse solutions or improve over time. This is consistent with prior observations that sparse rewards make it challenging for RL to associate actions with rewards~\citep{LADOSZ20221}.

In contrast, \textsc{CoCoS} employs the same RLOO training algorithm~\citep{ahmadian-etal-2024-back} as SCoRe but benefits from a progressive reward scheme. Under this setting, correctly solving even one more test case in a subsequent turn produces a positive reward, whereas making additional mistakes leads to a negative reward. This dynamic feedback encourages the expansion of the exploration space during training, enabling a more stable learning process. This progressive reward scheme encourages the expansion of the exploration space during training, enabling a more stable learning process.

\section{Additional Case Studies}\label{apdx:case_study}
We present two additional case studies that illustrate distinct types of errors: (2) complex logic errors and (3) incorrect usage of code libraries. These additional cases serve to complement the discussion in the main text by illustrating error types that extend beyond naming issues. They underscore \textsc{CoCoS}’s capacity to detect and remediate a diverse set of programming errors.

\begin{lstlisting}[language=Python,linewidth=\columnwidth]
# Initial response
def amicable_numbers_sum(n):
    sum = 0
    for i in range(1, n+1):
        # just adds divisor sums
        sum += sum_of_divisors(i)
    return sum

def sum_of_divisors(num):
    divisors = [1]
    for i in range(2, int(num**0.5) + 1):
        if num % i == 0:
            divisors.append(i)
            if i!= num // i:
                divisors.append(num // i)
    return sum(divisors)
\end{lstlisting}
\vspace{-1mm}
\begin{lstlisting}[language=Python,backgroundcolor=\color{ll1},caption=Example of resolving a multi-function interaction error,label=code:complex,linewidth=\columnwidth]
# Corrected response
def sum_of_divisors(n):
    divisors = [1]
    for i in range(2, int(n**0.5) + 1):
        if n % i == 0:
            divisors.append(i)
            if i!= n // i:
                divisors.append(n // i)
    return sum(divisors)

def amicable_numbers_sum(n):
    sum = 0
    for i in range(1, n+1):
        div_sum = sum_of_divisors(i)
        # check if i and div_sum form an amicable pair
        if div_sum != i and sum_of_divisors(div_sum) == i:
            sum += i
    return sum
\end{lstlisting}
Code~\ref{code:complex} illustrates an error involving the coordination of multiple functions. The model initially fails to implement the correct interaction logic across functions, resulting in an incorrect solution. In the subsequent turn, it is able to revise the necessary parts while preserving the correct subcomponents, effectively resolving the issue.

\begin{lstlisting}[language=Python,linewidth=\columnwidth]
# Initial response
def remove_dirty_chars(s1, s2):
    # Invalid in Python 3
    return s1.translate(None, s2)
\end{lstlisting}
\vspace{-1mm}
\begin{lstlisting}[language=Python,backgroundcolor=\color{ll1},caption=Example of using a non-existent method, label=code:method,linewidth=\columnwidth]
# Corrected response
def remove_dirty_chars(s1, s2):
    result = ""
    for char in s1:
        if char not in s2:
            result += char
    return result
\end{lstlisting}
Lastly, Code~\ref{code:method} demonstrates a misuse of a Python method. The model generates an invalid call to \texttt{str.translate()} that is not supported in the current version of Python. In the subsequent turn, it corrects the implementation by replacing the unsupported method with a valid character filtering loop.
\clearpage
\clearpage
\section{Prompts}\label{apdx:prompts}
In this section, we present the prompts used for prompt-based approaches and for \textsc{CoCoS} training and evaluation. The prompts for implementing prompting-based approaches are described in \S\ref{prompt:self-refine} and \S\ref{prompt:sets}. Finally, the prompts used for \textsc{CoCoS} training and evaluation are reported in \S\ref{prompt:main}.

\subsection{Main experiments}\label{prompt:main}
We present the instructions for evaluating the MBPP, HumanEval, and datasets. Tokens such as \texttt{[BEGIN]}, \texttt{[DONE]}, and \texttt{[CORRECT]} were used to delimit model outputs, and the prompts were created from \citet{kumar2025training}.

\begin{figure}[h]
\begin{prompt}[title={Prompt \thetcbcounter: Prompt used for generating initial code responses in the first turn}, label=fig:prompt_main_first]
You are an expert Python programmer, and here is your task: \{\{problem\}\} Your code should pass these tests:\\

\{\{test cases\}\}\\

[BEGIN]
\end{prompt}
\end{figure}

\begin{figure}[h]
\begin{prompt}[title={Prompt \thetcbcounter: Prompt used for generating self-corrected responses}, label=fig:prompt_main_second]
You are an expert Python programmer, and here is your task: \{\{problem\}\} Your code should pass these tests:\\

\{\{test cases\}\}\\

[BEGIN]

\{\{initial code\}\}

[DONE]\\

There might be an error in the code above because of a lack of understanding of the question. Please correct the error, if any, and rewrite the solution. Only output the final correct Python program!\\

[CORRECT]
\end{prompt}
\end{figure}

\clearpage
\subsection{Self-Refine}\label{prompt:self-refine}
The prompts were designed and implemented based on the manuscript, with Self-Refine~\citep{NEURIPS2023_91edff07} using a 3-shot prompting for both feedback generation and correction. Additionally, as the implementation was conducted without further fine-tuning, backticks such as \verb|```python| and \verb|```| were used to delimit model outputs.

\begin{figure}[h]
\begin{prompt}[title={Prompt \thetcbcounter: Prompt used for generating feedback in Self-Refine}]
I have some code. Can you give one suggestion to improve the solution to the problem. Don't fix the code, just give one suggestion.\\

Problem:

\{\{problem\}\}\\

Code:

\verb|```python|

\{\{initial code\}\}

\verb|```|\\

Feedback:
\end{prompt}
\end{figure}

\begin{figure}[h]
\begin{prompt}[title={Prompt \thetcbcounter: Prompt used for self-correction in Self-Refine}]
I have some code. Can you give one suggestion to improve the solution to the problem. Don't fix the code, just give one suggestion.\\

Problem:

\{\{problem\}\}\\

Code:

\verb|```python|

\{\{initial code\}\}

\verb|```|\\

Feedback:

\{\{feedback\}\}\\

Now fix the code.\\

Fixed Code:

\verb|```python|
\end{prompt}
\end{figure}

\clearpage
\subsection{SETS}\label{prompt:sets}
SETS~\citep{chen2025sets} employs test-time scaling laws~\citep{snell2025scaling} for self-correction, which precludes direct evaluation of Accuracy@t1 in \S\ref{sec:prompting-based}. Accordingly, we approximate it by reporting the average accuracy of the sampled initial responses. We use a sampling temperature of 0.7 and set the number of generation to 20.
To distinguish Python code from instructions, we used backticks, while the remaining instructions were created based on the manuscript.

\begin{figure}[h]
\begin{prompt}[title={Prompt \thetcbcounter: Prompt used for verifier in SETS}]
You are an expert in solving coding problems. You are given a PROBLEM and a PROPOSED CODE. Your job is to:

1. Transform the PROPOSED CODE into a statement given the PROBLEM and identify all constraints in the PROBLEM for verifying the statement.

2. Think step by step to verify if the statement satisfies each of the constraints.

3. Write a line of the form "The statement is correct" or "The statement is incorrect" at the end of your response based on your analysis.\\

PROBLEM:

\{\{problem\}\}\\

PROPOSED CODE:

\verb|```python|

\{\{initial code\}\}

\verb|```|\\

ANALYSIS:
\end{prompt}
\end{figure}

\begin{figure}[h]
\begin{prompt}[title={Prompt \thetcbcounter: Prompt used for self-correction in SETS}]
You are an expert in solving coding problems. You are given a PROBLEM and a set of CODE-ANALYSIS pairs. Your job is to generate a correct answer.\\

PROBLEM:

\{\{problem\}\}\\

CODE:

\verb|```python|

\{\{initial code\}\}

\verb|```|\\

ANALYSIS:

\{\{analysis\}\}\\

CORRECTED CODE:

\verb|```python|
\end{prompt}
\end{figure}

\clearpage
\section{Auxiliary instruction}\label{prompt:auxiliary instruction}

\begin{figure}[h]
\begin{prompt}[title={Prompt \thetcbcounter: Templates for the five rephrased instruction prompts.}]
Original: \\
There might be an error in the code above because of a lack of understanding of the question. Please correct the error, if any, and rewrite the solution. Only output the final correct Python program!\\

Insturction 1:\\
There might be an error in the code above because of a lack of understanding of the question. Please correct the error, if any, and rewrite the solution. Only output the final correct Python program!\\

Instruction 2:\\
If the solution above contains any mistakes due to a misunderstanding of the problem, fix them and rewrite the code. Only return the corrected Python program.\\

Instruction 3:\\
Check the code for potential errors caused by misinterpreting the task. Correct any issues and output just the final Python solution.\\

Instruction 4:\\
Review the code for possible bugs or logic errors. If needed, fix them and provide only the updated Python code as your answer.\\

Instruction 5:\\
Make sure the code fully matches the problem description. If any part is incorrect, fix it and return just the corrected Python code.
\end{prompt}
\end{figure}

\end{document}